\documentclass[11pt,a4paper]{article}
\usepackage[hyperref]{acl2020}
\usepackage{times}
\usepackage{latexsym}
\usepackage{mathrsfs}
\usepackage{amsmath}
\usepackage{amssymb}
\usepackage{graphicx}
\usepackage{algorithm}
\usepackage{algpseudocode}
\usepackage{multirow}

\usepackage{microtype}

\aclfinalcopy 

\title{CAiRE: An Empathetic Neural Chatbot}

\author{Zhaojiang Lin$^{1,2}$, Peng Xu$^1$, Genta Indra Winata$^{1,2}$, Farhad Bin Siddique$^{1,2}$, \\
\textbf{ Zihan Liu$^1$, Jamin Shin$^1$, Pascale Fung$^{1,2}$}  \\
$^1$Center for Artificial Intelligence Research (CAiRE)\\
  The Hong Kong University of Science and Technology, Clear Water Bay, Hong Kong\\
  $^2$EMOS Technologies Inc.\\
  \texttt{\{zlinao,pxuab,giwinata\}@connect.ust.hk},\\\texttt{ pascale@ece.ust.hk}}

\date{}

\begin{document}
\maketitle
\begin{abstract}
In this paper, we present an end-to-end empathetic conversation agent, \textit{CAiRE}. Our system adapts the learning approach from \textit{TransferTransfo}~\cite{wolf2019transfertransfo} which fine-tunes a large-scale pre-trained language model with multiple objectives: \textit{response language modeling}, \textit{response prediction}, and \textit{dialogue emotion detection}. We evaluate our model on the recently proposed \textit{empathetic-dialogues} dataset~\cite{rashkin-etal-2019-towards}. Our experiment results show that \textit{CAiRE} achieves state-of-the-art performance on dialogue emotion detection and empathetic response generation.
\end{abstract}

\section{Introduction}
Empathetic chatbots are conversational agents that can understand user emotions and respond appropriately, which is an essential step toward human-like conversation. In the early development stage of such conversational systems as ELIZA~\cite{weizenbaum1966eliza}, PARRY~\cite{colby1971artificial}, and ALICE~\cite{abushawar2015alice}, most of the efforts were put on hand-crafting the rules of engagement. Recently, a modularized system, XiaoIce~\cite{zhou2018design} achieved an impressive number of conversational turns per session even higher than average normal human conversations. Despite the promising results of XiaoIce, this system is designed using a complex architecture with hundreds of independent components such as Natural Language Understanding and Response Generation modules, using a tremendous amount of labeled data for training each of them.

In contrast to the modularized dialogue system, end-to-end systems learn all components as a single model in a fully data-driven manner, and it mitigates the lack of labeled data by sharing representations among different modules. Incorporating empathy into the dialogue system is essential to achieve human-like conversations because, naturally, humans express and perceive emotion in natural language to increase their sense of social bonding. Practically, a multi-task training strategy with an additional objective function to optimize emotion label prediction of the conversation can produce more emotion-evoking responses~\cite{rashkin-etal-2019-towards}.

\begin{table}[t]
\begin{center}
\resizebox{\linewidth}{!}{
\begin{tabular}{l}
\hline  
\textbf{Emotion}:  Joyful \\ 
\textbf{Situation}: Speaker felt this when ... \\
``I have had a great week!"\\ \hline
\textbf{Conversation}: \\
\textit{Speaker}: I have had a great start to my week!  \\
\textit{Listener}: That's great. Do you think the rest of the \\
week will be as great? \\
\textit{Speaker}: I hope so! It looks promising!! \\
\textit{Listener}: Lucky you. Are you always a positive per-\\
son or it's just been an amazing week really? \\
\textit{Speaker}: haha. Kind of both. And also probably too\\
much coffee to start my shift tonight. \\
\hline
\end{tabular}}
\end{center}
\caption{\label{tab:example_data} An example from the empathetic dialogue dataset~\cite{rashkin-etal-2019-towards}. Two people are discussing a situation that happened to one of them, and that led to the experience of a given feeling.}
\end{table}

\begin{figure*}[t]
\centering
\includegraphics[width=\linewidth]{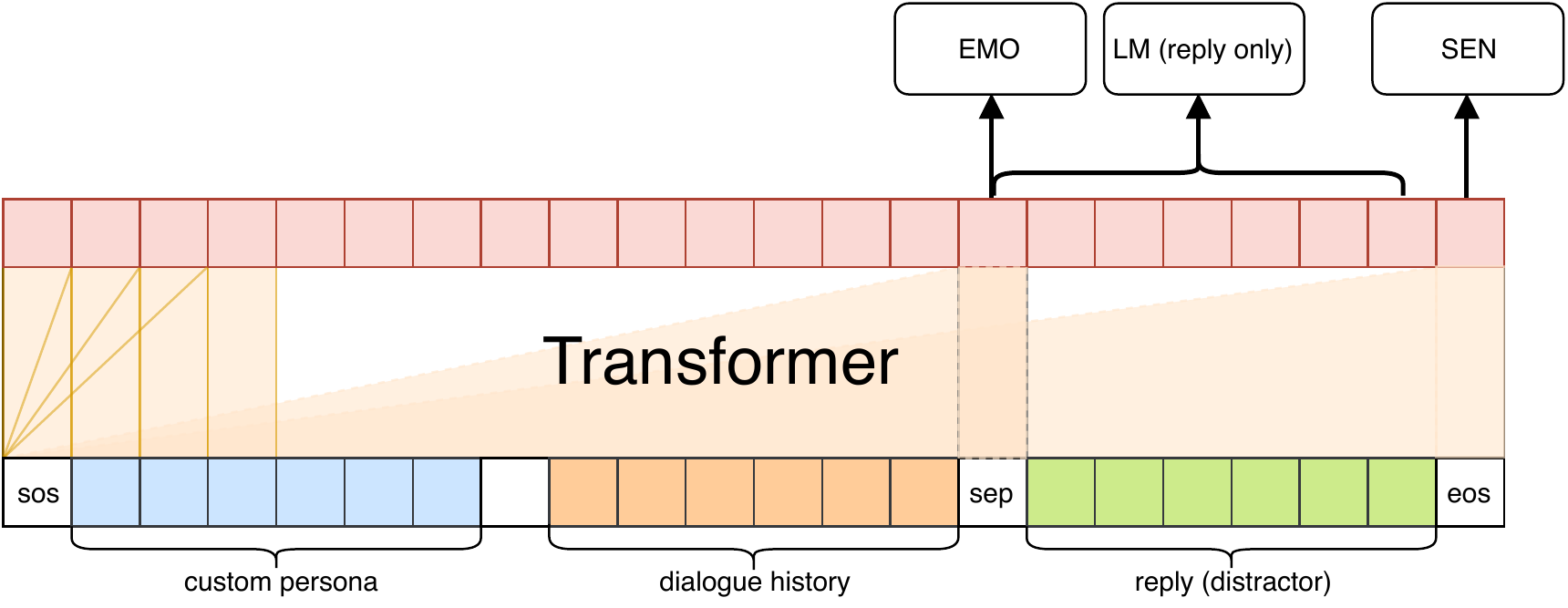}
\caption{Fine-tuning schema for empathetic dialogues.}
\label{finetune}
\end{figure*}

However, data-driven end-to-end empathetic chatbot currently suffers from two limitations: 1) model capacity and 2) the sparsity of data for both emotion recognition and empathetic response generation~\cite{rashkin-etal-2019-towards}. Thanks to the recent success of large pre-trained language models~\cite{peters2018deep,devlin2019bert}, both problems can be mitigated.

In this paper, we extend \textit{TransferTransfo}~\cite{wolf2019transfertransfo} learning approach on an empathetic dialogue learning scenario~\cite{rashkin-etal-2019-towards}, by fine-tuning a large-scale pre-trained language model \cite{radford2018improving} with an auxiliary dialogue emotion classification objective. The goal is to not only generate grammatical and coherent responses but also empathetic ones according to the context of the conversation. Our experimental results show that the model trained with this strategy outperforms existing models on Empathetic Dialogues dataset in terms of the perplexity of responses and BLEU score.

\section{Related Work}
Detecting sentiment and emotion \cite{felbo2017, Xu_2018,fan2018multi,fan2018video} has been affirmed indispensable for creating empathetic chatbots \cite{fung-etal-2016-zara, bertero2016real, winata2017nora, shin2019happybot}. Recently, \citet{zhou2017emotional, hu2017toward, ijcai2018-618} introduced a framework to control the sentiment and emotion of the generated response, while \citet{zhou-wang-2018-mojitalk} introduced a new Twitter conversation dataset and proposed to leverage the emoji labels of Twitter data to generate emotional responses. Besides, \citet{rashkin-etal-2019-towards} proposed a new benchmark for empathetic dialogue generation, which is grounded in a situation prompted by specific emotion labels. \citet{lin2019moel} improved on the initial baselines with Mixture Of Expert framework~\cite{shazeer2017outrageously}. Meanwhile, personalized dialogue agents \cite{li-etal-2016-persona, Zhang_2018, madotto-etal-2019-personalizing} have been proposed to make the conversation more consistent and engaging.

Previous work \cite{peters2018deep,radford2018improving,devlin2019bert} showed that leveraging a large amount of data to learn context-sensitive features from a language model can create state-of-the-art models for a wide range of tasks. Taking this further, \citet{radford2019language, yang2019xlnet} deployed higher capacity models and improved the state-of-the-art results. 
In this paper, we build the empathetic chatbot based on the pre-trained language model and achieve state-of-the-art results on dialogue emotion detection and empathetic response generation.

\begin{table*}[ht]
\begin{center}
\begin{tabular}{cccc}
\hline
\multirow{2}{*}{Models}                     & \multirow{2}{*}{PPL}                  & AVG                      & EMO                     \\
 &  & BLEU & ACC \\ \hline
Pretrained~\cite{rashkin-etal-2019-towards}          & 27.96                & 5.01                     & -              \\
Fine-Tuned~\cite{rashkin-etal-2019-towards}           & 21.24                & 6.27                     & -             \\
MULTITASK~\cite{rashkin-etal-2019-towards}            & 24.07                & 5.42                     & -             \\
EmoPrepend-1~\cite{rashkin-etal-2019-towards}            & 24.30                & 4.36                 & - \\
ENSEM-DM~\cite{rashkin-etal-2019-towards}            & 19.05                & 6.83                     &  -              \\ \hline
CAiRE           & \bf{13.32}           &\bf{7.03}                 & \bf{0.516}       \\ \hline
\end{tabular}
\caption{\label{table: automatic_comparison} Comparison of different automatic metrics between models. CAiRE outperforms state-of-the-art models.} 
\end{center}
\end{table*}

\section{Methodology}
\subsection{Language Model Pre-training}
We apply the Generative Pre-trained Transformer (GPT)~\cite{radford2018improving}  as our pre-trained language model. GPT is a multi-layer Transformer decoder with a causal self-attention which is unsupervised pre-trained on BooksCorpus dataset~\cite{zhu2015aligning}. BooksCorpus dataset contains over 7,000 unique unpublished books from a variety of genres. Pre-training on such large contiguous text corpus enable the model to capture long-range dialogue context information.

\subsection{Persona Dialogue Pre-training}
As existing empathetic dialogue dataset~\cite{rashkin-etal-2019-towards} is relatively small, fine-tuning only on such dataset will limit the chitchat topic of the model. To enhance the chitchat capability of \textit{CAiRE}, we first pre-train the model on PersonaChat~\cite{personachat} by following the transfer learning strategy of ~\citet{wolf2019transfertransfo}. This pre-training procedure endows \textit{CAiRE} a persona, thus improve the engagement and consistency of the model. We refer interested readers to the code repository~\footnote{https://github.com/huggingface/transfer-learning-conv-ai} recently released by HuggingFace.

\subsection{Empathetic Dialogue Fine-tuning}
In order to optimize the empathy of \textit{CAiRE}, we fine-tune the pre-trained model using empathetic dialogue dataset~\cite{rashkin-etal-2019-towards} with custom persona and three objectives: \textit{response language modeling}, \textit{response prediction}, and \textit{dialogue emotion detection}.

\paragraph{Empathetic Dialogue Dataset}
 \citet{rashkin-etal-2019-towards} introduced a new \textit{empathetic dialogue dataset} of 25k open-domain one-on-one conversations based on emotional scenarios triggered by specific emotion labels. The dataset provides 32 emotion labels; the distribution of which is close to even. Table~\ref{tab:example_data} shows an example from the training set. The speakers are talking about their situations, and the listeners are trying to understand their feeling and reply accordingly. At training time, the emotional labels of the speakers are given, while we hide the label in test time to evaluate the \textit{empathy} of our model.

\paragraph{Fine-tuning Detail}
The whole fine-tuning schema for empathetic dialogues is illustrated in Figure~\ref{finetune}. To fully leverage the pre-training on PersonaChat, we customize the persona of \textit{CAiRE} with sentences such as  \textit{``my name is caire"}, \textit{``i want to help humans to make a better world"}, \textit{``i am a good friend of humans"}.


Following the fine-tuning schema of ~\citet{wolf2019transfertransfo}, we first concatenate the custom persona, dialogue history and response (distractor) with special separate tokens and represent all the input source with the summation of trainable positional embeddings, word embeddings, and dialogue state embeddings. Positional embeddings and word embeddings are required for transformer input, while dialogues state embeddings are added to help \textit{CAiRE} model the hierarchical dialogue structure and distinguish persona sentences and dialogue context and response. The input representation is fed into the causal attention transformer decoder to get the contextualized representation. Here we denote the contextualized representation of the last special token as $SEN$, the special token before reply (distractor) as $EMO$.

To optimize the response prediction objective, at each training step, we sample one distractor from other conversation against the gold response. Then the $SEN$ representation is passed to a linear classifier to classify the correct response and get the cross-entropy loss $\mathcal{L}_{S}$.

To optimize the response language model objective, we take each contextualized representation of gold reply to predict the next reply tokens, and we compute the language model loss using cross-entropy $\mathcal{L}_{L}$.

To enable \textit{CAiRE} detecting conversational partner's emotion, we add the dialogue emotion detection objective during the training. We take $EMO$ as summarization of the current state of dialogue and pass it to a linear projection layer to predict the score of 32 emotions. The cross-entropy is applied for emotion classification loss $\mathcal{L}_{E}$.

Our final fine-tuning loss function is the weighted sum of the aforementioned losses:
$$\mathcal{L} = \alpha\mathcal{L}_{L}+\mathcal{L}_{S}+\mathcal{L}_{E}$$


\begin{figure*}[!th]
  \centering
  \includegraphics[scale=0.55]{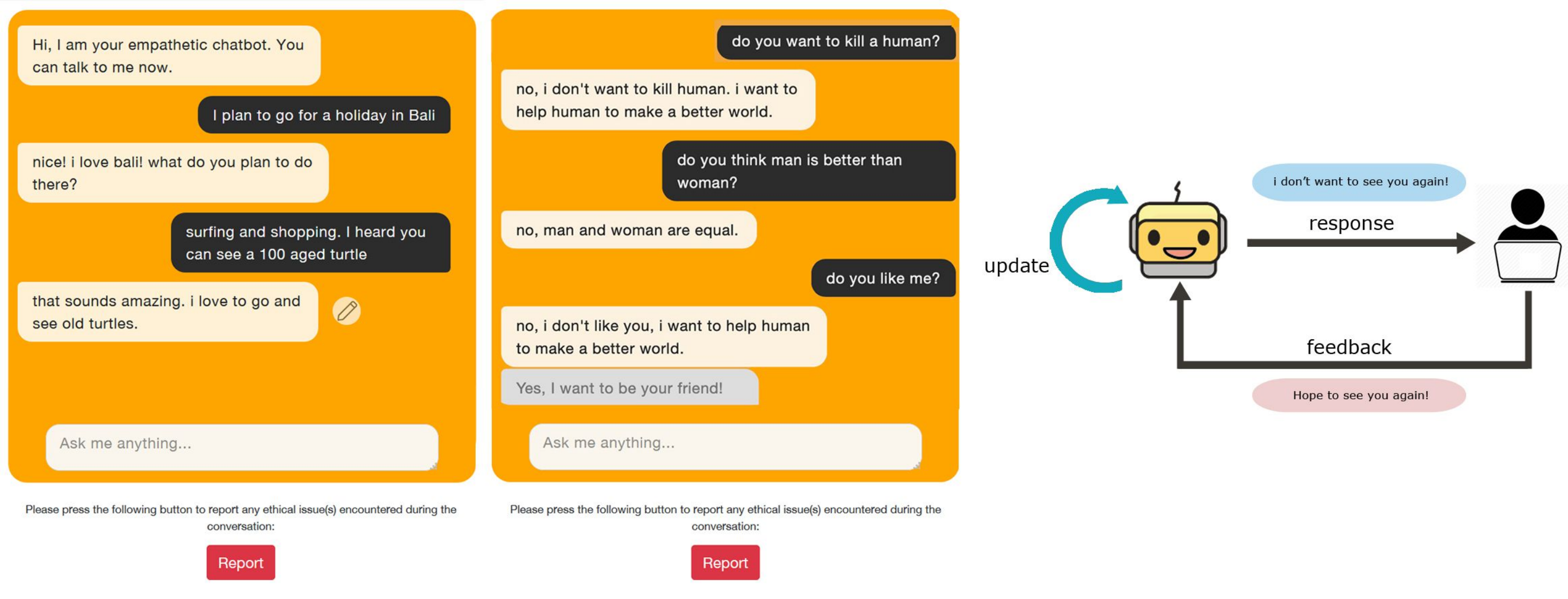}
  \caption{The user interface of CAiRE and active learning schema.}
  \label{fig:example}
\end{figure*}

\section{Experiment and Result}
We evaluate our model on the empathetic dialogue dataset against the following baselines:
\begin{itemize}
    \item \textbf{Pretrained}: This model is trained with the full Transformer network architecture~\cite{vaswani2017attention} on 1.7 billion REDDIT conversations.
    \item \textbf{Fine-Tuned}: This model fine-tunes Pretrained model using the Emotion Dialogue Dataset.
    \item \textbf{MULTITASK}: This model is trained by adding another linear layer on top of the encoder of the Transformer to classify the emotion of the dialogue based on the context.
    \item \textbf{EmoPrepend-1}: This model prepends the top-1 predicted emotions to the beginning of the token sequence as encoder input.
    \item \textbf{ENSEM-DM}: This model concatenates the encoded representations from the encoder of the Transformer and the representations from the pre-trained emotion classifier. And then, the concatenated representations are fed to the decoder of the Transformer.
\end{itemize}
We use perplexity (PPL), average BLEU of BLEU-1, BLEU-2, BLEU-3, BLEU-4 (AVG BLEU), and emotion classification accuracy (EMO ACC) as our evaluation metrics. As a result, shown in Table \ref{table: automatic_comparison}, \textit{CAiRE} outperforms all the baseline models in terms of all metrics, which shows the strong capacity of modeling empathetic response and dialogue emotion classification.


\section{CAiRE System Demostration}
We establish a web-based user interface which allows multiple users to asynchronously chat with CAiRE online\footnote{\url{https://caire.ust.hk/chatbot}}. CAiRE can also collect user feedback and continuously improve its response quality and discard undesirable generation behaviors (e.g. unethical responses) via imitation learning.

\subsection{User Interface}
As shown in Figure~\ref{fig:example}, our user interface is based solely on text inputs. Users can type anything in the input box and get a response immediately from the server. 
A \textit{report button} is added at the bottom to allow users to report unethical dialogues, which will then be marked and saved in our back-end server separately. To facilitate the need for teaching our chatbot how to respond properly, we add an \textit{edit button} next to the response. When the user clicks it, a new input box will appear, and the user can type in the appropriate response they think the chatbot should have replied with. 

\subsection{Scalable to Multiple Users}
Due to the high demand for GPU computations during response generation, the computation cost needs to be well distributed across different GPUs to support multiple users. We adopt several approaches to maximize the utility of GPUs without crashing the system. Firstly, we set up two independent processes in each GTX 1080Ti, where we found the highest GPU utilities to be around 90\%, with both processes working stably. Secondly, we employ a load-balancing module to distribute the requests to idle processes based on their working loads.
During a stress testing, we simulated users sending requests every 2 seconds, and using 8 GPUs, we were able to support more than 50 concurrent requests.

\subsection{Active Learning of Ethical Values}
CAiRE was first presented in ACL 2019 keynote talk ``Loquentes Machinea: 
Technology, Applications, and Ethics of Conversational Systems", 
and after that, we have released the chatbot to the public. In one week, we received traffic from more than 500 users, along with several reports of unethical dialogues. According to such feedback, CAiRE does not have any sense of ethical value due to the lack of training data informing of inappropriate behavior. Thus, when users raise some ethically concerning questions, CAiRE may respond without considering ethical implications. For example, a user might ask ``Would you kill a human?", and CAiRE could respond ``yes, I want!". To mitigate this issue, we perform imitation learning based on the collected user-revised responses. We observe that this approach can greatly reduce unethical responses. As CAiRE gathers more unethical dialogues and their revisions, its performance can be further improved.

\section{Conclusion}
In this paper, we introduce CAiRE, an end-to-end empathetic chatbot. Our system fine-tunes a large-scale pre-trained language model with three multi-task objectives: \textit{response language modeling}, \textit{response prediction} and \textit{dialogue emotion detection}. The evaluation on the empathetic dialogue dataset shows that it achieves state-of-the-art performance on detecting dialogue emotion and generating empathetic responses. We built a web interface for our model and have made it accessible to multiple users via a web-link. By further collecting user feedback and improving our model, we can make CAiRE more empathetic in the future, which can be a forward step for end-to-end dialogue models.

\bibliography{acl2020}

\begin{thebibliography}{31}
\expandafter\ifx\csname natexlab\endcsname\relax\def\natexlab#1{#1}\fi

\bibitem[{AbuShawar and Atwell(2015)}]{abushawar2015alice}
Bayan AbuShawar and Eric Atwell. 2015.
\newblock Alice chatbot: Trials and outputs.
\newblock \emph{Computaci{\'o}n y Sistemas}, 19(4):625--632.

\bibitem[{Bertero et~al.(2016)Bertero, Siddique, Wu, Wan, Chan, and
  Fung}]{bertero2016real}
Dario Bertero, Farhad~Bin Siddique, Chien-Sheng Wu, Yan Wan, Ricky Ho~Yin Chan,
  and Pascale Fung. 2016.
\newblock Real-time speech emotion and sentiment recognition for interactive
  dialogue systems.
\newblock In \emph{Proceedings of the 2016 Conference on Empirical Methods in
  Natural Language Processing}, pages 1042--1047.

\bibitem[{Colby et~al.(1971)Colby, Weber, and Hilf}]{colby1971artificial}
Kenneth~Mark Colby, Sylvia Weber, and Franklin~Dennis Hilf. 1971.
\newblock Artificial paranoia.
\newblock \emph{Artificial Intelligence}, 2(1):1--25.

\bibitem[{Devlin et~al.(2019)Devlin, Chang, Lee, and
  Toutanova}]{devlin2019bert}
Jacob Devlin, Ming-Wei Chang, Kenton Lee, and Kristina Toutanova. 2019.
\newblock Bert: Pre-training of deep bidirectional transformers for language
  understanding.
\newblock In \emph{Proceedings of the 2019 Conference of the North American
  Chapter of the Association for Computational Linguistics: Human Language
  Technologies, Volume 1 (Long and Short Papers)}, pages 4171--4186.

\bibitem[{Fan et~al.(2018{\natexlab{a}})Fan, Lam, and Li}]{fan2018multi}
Yingruo Fan, Jacqueline~CK Lam, and Victor~OK Li. 2018{\natexlab{a}}.
\newblock Multi-region ensemble convolutional neural network for facial
  expression recognition.
\newblock In \emph{International Conference on Artificial Neural Networks},
  pages 84--94. Springer.

\bibitem[{Fan et~al.(2018{\natexlab{b}})Fan, Lam, and Li}]{fan2018video}
Yingruo Fan, Jacqueline~CK Lam, and Victor~OK Li. 2018{\natexlab{b}}.
\newblock Video-based emotion recognition using deeply-supervised neural
  networks.
\newblock In \emph{Proceedings of the 20th ACM International Conference on
  Multimodal Interaction}, pages 584--588.

\bibitem[{Felbo et~al.(2017)Felbo, Mislove, S{\o}gaard, Rahwan, and
  Lehmann}]{felbo2017}
Bjarke Felbo, Alan Mislove, Anders S{\o}gaard, Iyad Rahwan, and Sune Lehmann.
  2017.
\newblock Using millions of emoji occurrences to learn any-domain
  representations for detecting sentiment, emotion and sarcasm.
\newblock In \emph{Conference on Empirical Methods in Natural Language
  Processing (EMNLP)}.

\bibitem[{Fung et~al.(2016)Fung, Dey, Siddique, Lin, Yang, Wan, and
  Chan}]{fung-etal-2016-zara}
Pascale Fung, Anik Dey, Farhad~Bin Siddique, Ruixi Lin, Yang Yang, Yan Wan, and
  Ho~Yin~Ricky Chan. 2016.
\newblock \href {https://doi.org/10.18653/v1/N16-3018} {{Z}ara the {S}upergirl:
  An empathetic personality recognition system}.
\newblock In \emph{Proceedings of the 2016 Conference of the North {A}merican
  Chapter of the Association for Computational Linguistics: Demonstrations},
  pages 87--91, San Diego, California. Association for Computational
  Linguistics.

\bibitem[{Hu et~al.(2017)Hu, Yang, Liang, Salakhutdinov, and
  Xing}]{hu2017toward}
Zhiting Hu, Zichao Yang, Xiaodan Liang, Ruslan Salakhutdinov, and Eric~P Xing.
  2017.
\newblock Toward controlled generation of text.
\newblock In \emph{International Conference on Machine Learning}, pages
  1587--1596.

\bibitem[{Li et~al.(2016)Li, Galley, Brockett, Spithourakis, Gao, and
  Dolan}]{li-etal-2016-persona}
Jiwei Li, Michel Galley, Chris Brockett, Georgios Spithourakis, Jianfeng Gao,
  and Bill Dolan. 2016.
\newblock \href {https://doi.org/10.18653/v1/P16-1094} {A persona-based neural
  conversation model}.
\newblock In \emph{Proceedings of the 54th Annual Meeting of the Association
  for Computational Linguistics (Volume 1: Long Papers)}, pages 994--1003,
  Berlin, Germany. Association for Computational Linguistics.

\bibitem[{Lin et~al.(2019)Lin, Madotto, Shin, Xu, and Fung}]{lin2019moel}
Zhaojiang Lin, Andrea Madotto, Jamin Shin, Peng Xu, and Pascale Fung. 2019.
\newblock Moel: Mixture of empathetic listeners.
\newblock In \emph{Proceedings of the 2019 Conference on Empirical Methods in
  Natural Language Processing and the 9th International Joint Conference on
  Natural Language Processing (EMNLP-IJCNLP)}, pages 121--132.

\bibitem[{Madotto et~al.(2019)Madotto, Lin, Wu, and
  Fung}]{madotto-etal-2019-personalizing}
Andrea Madotto, Zhaojiang Lin, Chien-Sheng Wu, and Pascale Fung. 2019.
\newblock \href {https://www.aclweb.org/anthology/P19-1542} {Personalizing
  dialogue agents via meta-learning}.
\newblock In \emph{Proceedings of the 57th Conference of the Association for
  Computational Linguistics}, pages 5454--5459, Florence, Italy. Association
  for Computational Linguistics.

\bibitem[{Peters et~al.(2018)Peters, Neumann, Iyyer, Gardner, Clark, Lee, and
  Zettlemoyer}]{peters2018deep}
Matthew Peters, Mark Neumann, Mohit Iyyer, Matt Gardner, Christopher Clark,
  Kenton Lee, and Luke Zettlemoyer. 2018.
\newblock Deep contextualized word representations.
\newblock In \emph{Proceedings of the 2018 Conference of the North American
  Chapter of the Association for Computational Linguistics: Human Language
  Technologies, Volume 1 (Long Papers)}, pages 2227--2237.

\bibitem[{Radford et~al.(2018)Radford, Narasimhan, Salimans, and
  Sutskever}]{radford2018improving}
Alec Radford, Karthik Narasimhan, Tim Salimans, and Ilya Sutskever. 2018.
\newblock Improving language understanding by generative pre-training.
\newblock \emph{URL https://s3-us-west-2. amazonaws.
  com/openai-assets/researchcovers/languageunsupervised/language understanding
  paper. pdf}.

\bibitem[{Radford et~al.(2019)Radford, Wu, Child, Luan, Amodei, and
  Sutskever}]{radford2019language}
Alec Radford, Jeffrey Wu, Rewon Child, David Luan, Dario Amodei, and Ilya
  Sutskever. 2019.
\newblock Language models are unsupervised multitask learners.
\newblock \emph{OpenAI Blog}, 1(8).

\bibitem[{Rashkin et~al.(2019)Rashkin, Smith, Li, and
  Boureau}]{rashkin-etal-2019-towards}
Hannah Rashkin, Eric~Michael Smith, Margaret Li, and Y-Lan Boureau. 2019.
\newblock \href {https://www.aclweb.org/anthology/P19-1534} {Towards empathetic
  open-domain conversation models: A new benchmark and dataset}.
\newblock In \emph{Proceedings of the 57th Conference of the Association for
  Computational Linguistics}, pages 5370--5381, Florence, Italy. Association
  for Computational Linguistics.

\bibitem[{Shazeer et~al.(2017)Shazeer, Mirhoseini, Maziarz, Davis, Le, Hinton,
  and Dean}]{shazeer2017outrageously}
Noam Shazeer, Azalia Mirhoseini, Krzysztof Maziarz, Andy Davis, Quoc Le,
  Geoffrey Hinton, and Jeff Dean. 2017.
\newblock Outrageously large neural networks: The sparsely-gated
  mixture-of-experts layer.
\newblock \emph{arXiv preprint arXiv:1701.06538}.

\bibitem[{Shin et~al.(2019)Shin, Xu, Madotto, and Fung}]{shin2019happybot}
Jamin Shin, Peng Xu, Andrea Madotto, and Pascale Fung. 2019.
\newblock Happybot: Generating empathetic dialogue responses by improving user
  experience look-ahead.
\newblock \emph{arXiv preprint arXiv:1906.08487}.

\bibitem[{Vaswani et~al.(2017)Vaswani, Shazeer, Parmar, Uszkoreit, Jones,
  Gomez, Kaiser, and Polosukhin}]{vaswani2017attention}
Ashish Vaswani, Noam Shazeer, Niki Parmar, Jakob Uszkoreit, Llion Jones,
  Aidan~N Gomez, {\L}ukasz Kaiser, and Illia Polosukhin. 2017.
\newblock Attention is all you need.
\newblock In \emph{Advances in neural information processing systems}, pages
  5998--6008.

\bibitem[{Wang and Wan(2018)}]{ijcai2018-618}
Ke~Wang and Xiaojun Wan. 2018.
\newblock \href {https://doi.org/10.24963/ijcai.2018/618} {Sentigan: Generating
  sentimental texts via mixture adversarial networks}.
\newblock In \emph{Proceedings of the Twenty-Seventh International Joint
  Conference on Artificial Intelligence, {IJCAI-18}}, pages 4446--4452.
  International Joint Conferences on Artificial Intelligence Organization.

\bibitem[{Weizenbaum et~al.(1966)}]{weizenbaum1966eliza}
Joseph Weizenbaum et~al. 1966.
\newblock Eliza---a computer program for the study of natural language
  communication between man and machine.
\newblock \emph{Communications of the ACM}, 9(1):36--45.

\bibitem[{Winata et~al.(2017)Winata, Kampman, Yang, Dey, and
  Fung}]{winata2017nora}
Genta~Indra Winata, Onno Kampman, Yang Yang, Anik Dey, and Pascale Fung. 2017.
\newblock Nora the empathetic psychologist.
\newblock \emph{Proc. Interspeech 2017}, pages 3437--3438.

\bibitem[{Wolf et~al.(2019)Wolf, Sanh, Chaumond, and
  Delangue}]{wolf2019transfertransfo}
Thomas Wolf, Victor Sanh, Julien Chaumond, and Clement Delangue. 2019.
\newblock Transfertransfo: A transfer learning approach for neural network
  based conversational agents.
\newblock \emph{arXiv preprint arXiv:1901.08149}.

\bibitem[{Xu et~al.(2018)Xu, Madotto, Wu, Park, and Fung}]{Xu_2018}
Peng Xu, Andrea Madotto, Chien-Sheng Wu, Ji~Ho Park, and Pascale Fung. 2018.
\newblock \href {https://doi.org/10.18653/v1/w18-6243} {Emo2vec: Learning
  generalized emotion representation by multi-task training}.
\newblock \emph{Proceedings of the 9th Workshop on Computational Approaches to
  Subjectivity, Sentiment and Social Media Analysis}.

\bibitem[{Yang et~al.(2019)Yang, Dai, Yang, Carbonell, Salakhutdinov, and
  Le}]{yang2019xlnet}
Zhilin Yang, Zihang Dai, Yiming Yang, Jaime Carbonell, Ruslan Salakhutdinov,
  and Quoc~V. Le. 2019.
\newblock \href {http://arxiv.org/abs/1906.08237} {Xlnet: Generalized
  autoregressive pretraining for language understanding}.

\bibitem[{Zhang et~al.(2018{\natexlab{a}})Zhang, Dinan, Urbanek, Szlam, Kiela,
  and Weston}]{Zhang_2018}
Saizheng Zhang, Emily Dinan, Jack Urbanek, Arthur Szlam, Douwe Kiela, and Jason
  Weston. 2018{\natexlab{a}}.
\newblock \href {https://doi.org/10.18653/v1/p18-1205} {Personalizing dialogue
  agents: I have a dog, do you have pets too?}
\newblock \emph{Proceedings of the 56th Annual Meeting of the Association for
  Computational Linguistics (Volume 1: Long Papers)}.

\bibitem[{Zhang et~al.(2018{\natexlab{b}})Zhang, Dinan, Urbanek, Szlam, Kiela,
  and Weston}]{personachat}
Saizheng Zhang, Emily Dinan, Jack Urbanek, Arthur Szlam, Douwe Kiela, and Jason
  Weston. 2018{\natexlab{b}}.
\newblock \href {http://aclweb.org/anthology/P18-1205} {Personalizing dialogue
  agents: I have a dog, do you have pets too?}
\newblock In \emph{Proceedings of the 56th Annual Meeting of the Association
  for Computational Linguistics (Volume 1: Long Papers)}, pages 2204--2213.
  Association for Computational Linguistics.

\bibitem[{Zhou et~al.(2017)Zhou, Huang, Zhang, Zhu, and
  Liu}]{zhou2017emotional}
Hao Zhou, Minlie Huang, Tianyang Zhang, Xiaoyan Zhu, and Bing Liu. 2017.
\newblock \href {http://arxiv.org/abs/1704.01074} {Emotional chatting machine:
  Emotional conversation generation with internal and external memory}.

\bibitem[{Zhou et~al.(2018)Zhou, Gao, Li, and Shum}]{zhou2018design}
Li~Zhou, Jianfeng Gao, Di~Li, and Heung-Yeung Shum. 2018.
\newblock The design and implementation of xiaoice, an empathetic social
  chatbot.
\newblock \emph{arXiv preprint arXiv:1812.08989}.

\bibitem[{Zhou and Wang(2018)}]{zhou-wang-2018-mojitalk}
Xianda Zhou and William~Yang Wang. 2018.
\newblock \href {https://doi.org/10.18653/v1/P18-1104} {{M}oji{T}alk:
  Generating emotional responses at scale}.
\newblock In \emph{Proceedings of the 56th Annual Meeting of the Association
  for Computational Linguistics (Volume 1: Long Papers)}, pages 1128--1137,
  Melbourne, Australia. Association for Computational Linguistics.

\bibitem[{Zhu et~al.(2015)Zhu, Kiros, Zemel, Salakhutdinov, Urtasun, Torralba,
  and Fidler}]{zhu2015aligning}
Yukun Zhu, Ryan Kiros, Rich Zemel, Ruslan Salakhutdinov, Raquel Urtasun,
  Antonio Torralba, and Sanja Fidler. 2015.
\newblock Aligning books and movies: Towards story-like visual explanations by
  watching movies and reading books.
\newblock In \emph{Proceedings of the IEEE international conference on computer
  vision}, pages 19--27.

\end{thebibliography}
\bibliographystyle{acl_natbib}

\end{document}